\documentclass{article}

\usepackage{amsmath} 
\DeclareMathOperator*{\argmin}{arg\,min}
\usepackage{graphicx} 
\usepackage{multirow} 
\usepackage{authblk} 

\begin{document}

\title{A Brief Introduction to the Temporal Group LASSO and its Potential 
  Applications in Healthcare}
\author[1]{Diego Saldana Miranda \thanks{diego.saldana\_miranda@novartis.com}}
\affil[1]{Novartis Pharma A.G., Real World Evidence IT, Basel, Switzerland}
\date{April 2017}
\maketitle

\begin{abstract}
The Temporal Group LASSO is an example of a multi-task, regularized 
regression approach for the prediction of response variables that vary over 
time. The aim of this work is to introduce the reader to the concepts behind the
Temporal Group LASSO and its related methods, as well as to the type of 
potential applications in a healthcare setting that the method has. We argue 
that the method is attractive because of its ability to reduce overfitting, 
select predictors, learn smooth effect patterns over time, and finally, its 
simplicity.
\end{abstract}

\section{Introduction}
The prediction of longitudinal, patient level outcomes has a wide range of 
potential applications in healthcare, including early safety signal detection 
\cite{safety_signal}, patient stratification \cite{patient_stratification}, 
assessment of difference of treatment effects \cite{effect_difference}  over 
time, prediction of symptom exhacerbations \cite{exhacerbations}  over time, 
prediction of patient cost blooming \cite{cost_blooming}, among many others.

There are many approaches to the modelling and prediction of longitudinal data, 
including mixed-effects regression models \cite{mixed_effects}, covariance 
pattern models \cite{covariance_pattern}, structural equations 
\cite{structural_equations} models, generalized estimating equations 
\cite{generalized_estimating_equations} models, as well as pharmacokinetics and 
pharmacodynamics based models \cite{pk_pd}, among others.

The Temporal Group LASSO was introduced by Zhou \emph{et al.} \cite{tg_lasso} 
as a method to predict longitudinal outcomes, select most relevant predictors 
(and simultaneously eliminate less relevant ones), as well as obtaining effect 
estimates for the selected variables that vary smoothly over time. 

This paper is organized as follows: in Section \ref{sec:preliminary_concepts} 
we introduce the reader to some of the concepts and methods that preceded the 
development of the Temporal Group LASSO; in Section \ref{sec:tg_lasso} we 
present the Temporal Group LASSO and its differences and similarities with 
traditional GLMs; in Section \ref{sec:brief_example} we present a brief 
example, applying the Temporal Group LASSO on a simulated clinical dataset; 
finally, we present our final thoughts and conclusions in Section 
\ref{sec:conclusion}.

\section{Preliminary Concepts}
\label{sec:preliminary_concepts}
Generalized linear models (GLMs) \cite{glm1} are widely used in healthcare for 
modelling outcomes and for estimating the effects that different explanatory 
variables have on them. GLMs generalize ordinary linear regression to non-linear
relationships by allowing the use of a \emph{link function} that relates the 
expected value of the response to the \emph{linear predictor}. When the link 
function is the identity, and the random component is the normal distribution, 
the GLM reduces to ordinary linear regression. In general, the problem of 
fitting a GLM can be written as 

\begin{equation}
\label{eq:glm}
  \mathbf{w}^{*} = \argmin_{\mathbf{w}}{L(\mathbf{w})}, 
\end{equation}

\noindent where the left side of the equation is the vector of coefficients that
will be assigned to the GLM. This set of coefficients may have different 
interpretations depending on the GLM model class, such as odds ratios in the 
case of logistic regression, hazard ratios in the case of a Cox proportional 
hazards model, etc. The right side of the equation represents the obtention of 
a set of coefficients that minimize the error of the GLM. The \emph{loss 
function}, $L(\mathbf{w})$ may be the least squares in the case of linear 
regression, the negative log-likelihood in the case of logistic regression, etc.
It is outside of the scope of this work to explain GLMs in detail, but the 
reader can refer to the large amount of existing literature on GLMs \cite{glm1, 
glm2, glm3}. 

Although GLMs are effective at modelling outcomes and effects, they
have some shortcomings in particular situations. For example, GLMs are prone to 
overfitting \cite{glm_limitations}, in particular when the number of predictors 
considered is very large \cite{glm_limitations}. A common approach in 
modern regression to reduce overfitting (at the expense of more 'shrinked' 
coefficients) is the use of ridge regression \cite{ridge_regression}. In ridge 
regression, the loss function in \ref{eq:glm} is modified through the addition 
of a penalty term, and the problem is now given by 

\begin{equation}
  \mathbf{w}^{*} = \argmin_{\mathbf{w}}{L(\mathbf{w}) + \lambda \| \mathbf{w}
    \|_{2}^{2}},
\end{equation}

\noindent where $\| \mathbf{w} \|_{2}^{2}$, called \emph{the squared $L_{2}$ 
norm of $\mathbf{w}$}, is equivalent to the sum of the squared coefficients in 
$\mathbf{w}$. Essentially, this means that the solution should minimize the 
error but also favor solutions with smaller coefficients (closer to zero), and 
the degree of tradeoff between smaller error and smaller coefficients is 
determined by the \emph{tuning} parameter $\lambda$, which is often chosen 
through cross validation.

Another shortcoming of GLMs is that they don't allow the selection of 
relevant predictors, so predictors need to be selected in some other way, 
for example using time and computationally consuming processes, such as subset 
selection \cite{glm_subset_stepwise} or step-wise selection 
\cite{glm_subset_stepwise}. Although ridge regression reduces overfitting and 
shrinks the coefficients by taking them closer to zero, it doesn't eliminate 
coefficients, and instead produces a solution with very small coefficients. A 
common approach from statistical learning to reduce overfitting as well as to 
select predictors is the use of the LASSO \cite{lasso}. Similar to the ridge 
regression case, in the LASSO, the loss function in \ref{eq:glm} is modified by 
the addition of a penalty term, and the problem given by 

\begin{equation}
  \mathbf{w}^{*} = \argmin_{\mathbf{w}}{L(\mathbf{w}) + \lambda \| \mathbf{w} 
    \|_{1},}
\end{equation} 

\noindent where $\| \mathbf{w} \|_{1}$ is known as \emph{the $L_{1}$ norm of 
$\mathbf{w}$}, and is equivalent to the sum of absolute values of the 
coefficients in $\mathbf{w}$. Like in ridge regression, this formulation favors 
solutions with 'shrinked' coefficients, but unlike in ridge regression, it 
favors solutions with zero valued coefficients as well, also referred to as 
'sparse' solutions. This means that the LASSO can be used to select relevant 
predictors, a task known as \emph{feature selection} in the machine learning 
literature.

Although both ridge regression and the LASSO enhance the power of GLMs, each one
of them has particular weaknesses, and while ridge regression doesn't yield 
sparse solutions, the LASSO lacks the ability to select multiple variables 
from a group of correlated variables, often selecting only one of them instead. 
An approach often used to combine the strengths of both ridge regression and the
LASSO is the Elastic Net \cite{elastic_net}. In Elastic Nets, the penalized 
regression problem is given by 

\begin{equation}
  \mathbf{w}^{*} = \argmin_{\mathbf{w}}{L(\mathbf{w}) + \lambda_{1} \| 
    \mathbf{w} \|_{1} + \lambda_{2} \| \mathbf{w} \|_{2}^{2}},
\end{equation}

\noindent with both the $L_{1}$ and squared $L_{2}$ penalties being applied 
jointly. Such a model is able to achieve both the sparse solutions needed for 
selecting predictors, as well as preventing the selection of only one variable 
from groups of highly correlated variables.

Finally, one limitation of the Elastic Net is that it is unable to select 
(and eliminate) groups of predictors. It would be useful for an algorithm to 
perform a group-wise selection, for example when having diverse sources of data,
such as laboratory data, vital signs, gene expression data, etc. One way to 
perform group-wise selection of predictors is the Group LASSO 
\cite{group_lasso}. In the group LASSO, the penalized regression problem is 
given by 

\begin{equation}
  \mathbf{w}^{*} = \argmin_{\mathbf{w}}{L(\mathbf{w}) + \lambda \sum_{g = 1}^{G}
    {\| \mathbf{w}^{(g)} \|_{2}}},
\end{equation}

\noindent where $G$ is the number of groups, and $\mathbf{w}^{(g)}$ is a vector 
containing the subset of elements in $\mathbf{w}$ that belong to the $g$th 
group. Notice that the $L_{2}$ norm is not squared, and in this case is 
$\| \mathbf{w}^{(g)} \|_{2}$ is equivalent to the square root of the sum of 
squared elements in $\mathbf{w}^{(g)}$. Notice also that in the particular case 
where there is one group $g$ for each element in $\mathbf{w}$ and having only 
that one element, this reduces to the LASSO.

\section{The Temporal Group LASSO}
\label{sec:tg_lasso}
The core of the Temporal Group LASSO \cite{tg_lasso} is a GLM with a 
multivariate response. In this case, the loss function can be written as 
$L(\mathbf{W})$, where $\mathbf{W}$ is a matrix of coefficients, rather than a 
vector. Each column of $\mathbf{W}$ corresponds to a specific time point in the 
data. For example, in the case of such a multivariate GLM with Gaussian error, 
the loss function would be given by 

\begin{equation}
  L(\mathbf{W}) = \| \mathbf{Y} - \mathbf{XW} \|_{F}^{2},
\end{equation}

\noindent where $\mathbf{Y}$ is a matrix of response values with one row per 
patient and one column per time point, $\mathbf{X}$ is a matrix of predictors 
with one row per patient and one column per predictor variable, note that we 
have omitted the intercepts for simplicty. The notation $\| \cdot \|_{F}^{2}$ 
simply stands for the sum of the squares of the elements of the matrix inside 
the norm, and it's referred to as \emph{the squared Frobenius norm}. This 
formulation of the loss is equivalent to the least squares loss that is normally
used in ordinary linear regression. Similar multivariate losses can be written 
for the other GLM classes.

This formulation has weaknesess similar to those of GLMs with a univariate 
response, such as susceptibility to overfitting, inability to perform feature 
selection, etc. In addition, there is no connection between the coefficients 
learned in different columns. Intuitively, a model where the coefficients for 
a given variable (the rows of $\mathbf{W}$ vary smoothly over time would be 
desirable.  Therefore, a model that allows the prevention of overfitting, 
the row-wise selection of predictors, as well as smooth variation over time is 
desired. The Temporal Group LASSO attempts to approach this using the following 
a formulation given by  

\begin{equation}
  \mathbf{W}^{*} = \argmin_{\mathbf{W}}{L(\mathbf{W}) + \lambda_{1} \| 
    \mathbf{W} \|_{F}^{2} + \lambda_{2} \| \mathbf{RW}^{T} \|_{F}^{2} + 
    \lambda_{3} \| \mathbf{W} \|_{2,1}}.
\end{equation}

This formulation has three penalties. The first one, $\| \mathbf{W} \|_{F}^{2}$ 
is equivalent to the sum of the squared coefficients in $\mathbf{W}$, which 
means that this penalty induces shrinkage of the coefficients in a similar way 
to ridge regression.

In the second penalty term, $\mathbf{R}$ is a matrix $T - 1 \times T$ matrix 
where $T$ is the number of time points, 

\begin{equation}
  R_{i,j} = 
  \begin{cases}
    i = j, 1 \\
    i + 1 = j, -1 \\
    otherwise, 0
  \end{cases},
\end{equation}

\noindent and this means that the formulation penalizes the squared differences 
of the coefficients associated to adjacent time points for each predictor. This 
can also be interpreted as a Laplacian term \cite{fsgl}, and essentially 
means that the Temporal Group LASSO favors solutions with coefficients that 
vary smoothly over time. In addition, it's possible to weight the differences 
using numbers other than one in order to penalize differences at different time 
points more than others. This may be desirable, for example, when measurements 
are taken at irregular intervals, or if a drastic change is actually expected at
a particular timepoint.

The third penalty term, $\| \mathbf{W} \|_{2,1}$, is called \emph{the $L_{2,1}$ 
norm of $\mathbf{W}$} and is equivalent to the sum of the $L_{2}$ norms of each 
of the rows of $\mathbf{W}$. This is equivalent to a Group LASSO penalty where 
the groups are the rows of $\mathbf{W}$, and it means that the formulation will 
favor solutions with row-wise sparsity. Since the rows of $\mathbf{W}$ are 
coefficients assigned to the same predictor but at different time points, this 
means that this penalty will favor the selection (and elimination) of the same 
predictors accross multiple time points. The selection of predictors across 
multiple tasks is known in the machine learning literature as \emph{joint 
feature selection}.

As a result of these three penalty terms, the Temporal Group LASSO is able to 
reduce overfitting, learn coefficients that vary smoothly over time, and perform
joint feature selection across time points. Finally, the model is simple to 
implement, with most of the penalties being smooth and differentiable, and the 
$L_{2,1}$ penalty being row-wise separable and for which the same methods used 
for other LASSO type methods, such as proximal gradient methods 
\cite{proximal_algorithms}, can be used.

Although Zhou \emph{et al.} introduced a stability sampling based method for 
feature selection, here we explore a cross validation based method, more 
similar to approaches used with other methods, such as the LASSO. Zhou later 
introduced other related models, such as the Fused Sparse Group LASSO
\cite{fsgl}. However, these models use somewhat more complex penalty schemes, 
and are outside of the scope of this paper.

\section{Brief Example on a Simulated Clinical Dataset}
\label{sec:brief_example}

In order to show an example of application of the Temporal Group LASSO, a 
simulated dataset has been generated based on a model that mimics the effect 
of 3 injections of warfarin over the course of 50 hours \cite{rx_ode}. Using 
this model, data for 500 patients has been simulated, of which 350 are used as a
training set, and 150 are used as a test set. A set of 300 randomly generated 
features following a gaussian distribution was generated for each patient, and 
the parameters of the simulation have been generated as linear functions of a 
subset of these features. In total, 170 features are involved in at least one of
the simulation parameters. Thus, the relationship between the features and the 
effect itself is generally not purely linear. Random noise has been added in 
order to account for random variation not related to any of the features in a 
dataset. In a real dataset, these features may be variables such as baseline 
gene expression \cite{gene_expression}, medical imaging features 
\cite{tg_lasso}, laboratory measurements, etc.

\begin{table}[h]
\centering
\caption{Summary characteristics and error metrics of the effect on the test 
  set.}
\begin{tabular}{ c | c | c | c | c | c }
  \hline\hline
   & & \multicolumn{2}{ c | }{TGL} & RR & MLR \\
  Hour & Mean ($\pm$ Std. Dev.) & RMSE & $R^2$ & $R^2$ & $R^2$ \\
  \hline
  5 & 0.500 ($\pm$ 0.058) & 0.025 &  0.810 & 0.715 & 0.382 \\
  10 & 0.278 ($\pm$ 0.059) & 0.028 & 0.773 & 0.691 & 0.405 \\
  15 & 0.241 ($\pm$ 0.059) & 0.030 & 0.749 & 0.666 & 0.446 \\
  20 & 0.275 ($\pm$ 0.069) & 0.034 & 0.753 & 0.647 & 0.413 \\
  25 & 0.287 ($\pm$ 0.074) & 0.036 & 0.761 & 0.649 & 0.408 \\
  30 & 0.165 ($\pm$ 0.053) & 0.027 & 0.733 & 0.639 & 0.395 \\
  35 & 0.134 ($\pm$ 0.043) & 0.022 & 0.723 & 0.633 & 0.391 \\
  40 & 0.155 ($\pm$ 0.044) & 0.023 & 0.739 & 0.635 & 0.387 \\
  45 & 0.192 ($\pm$ 0.052) & 0.025 & 0.756 & 0.638 & 0.362 \\
  50 & 0.178 ($\pm$ 0.050) & 0.024 & 0.758 & 0.647 & 0.384 \\
  \hline\hline
\end{tabular}
\label{tab:error_metrics}
\end{table}

After selection of the optimal regularization parameters through 
cross-validation, the final model selects 131 (77\%) of the 171 features that 
truly have an influence in the effect. The remaining features are not truly 
involved in the mechanism governing the effect. However, the predictions of 
the Temporal Group Lasso (TGL) model are highly accurate, as shown in Table 
\ref{tab:error_metrics}, and the model is able to explain more than 72\% of the 
variance at all time points in this dataset despite the presence of noise and 
the high amount of uninformative variables. Further, the model clearly 
outperforms Ridge Regression (RR) and Multiple Linear Regression (MLR).

\begin{figure}[h]
\centering
\includegraphics[scale=.3]{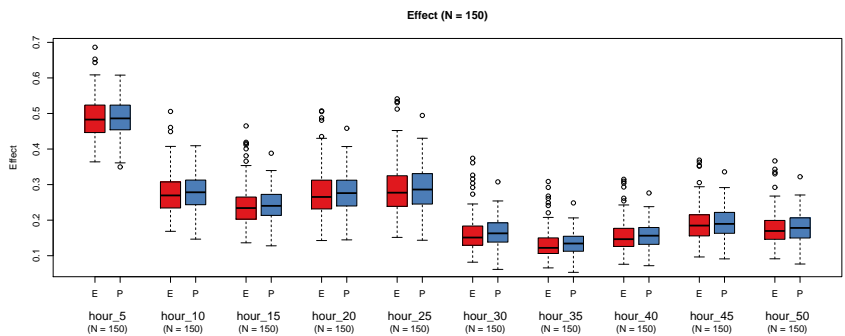}
\caption{Box plot of the effect at each time point for the expected (E), and 
  predicted (P) outcomes in the test set.}
\label{fig:boxplot}
\end{figure}

Figure \ref{fig:boxplot} shows the trends and spread of both the 
expected and predicted effect in the test set, which are highly similar. A 
number of outliers can be observed for the expected effect, though, which are 
not completely captured by the model. This may be due to the non-linear nature 
of the relationship between the features and the effect. An alternative may be 
to model the logarithm of the effect instead, however, we don't explore this 
alternative in this example.

\section{Conclusion}
\label{sec:conclusion}

We have presented here an introduction to the concepts behind the Temporal 
Group Lasso, a multi-task learning algorithm that can be used to predict 
outcomes as well as select important predictors while reducing over-fitting 
and achieving a smooth variation of the model coefficients over time points. 
We have also presented a brief example based on a simulation of a clinical 
treatment effect over the course of 50 hours in 500 patients. The model 
outperformed other linear regression methods, such as ridge regression and 
traditional multi-linear regression. 

The ability to select the most relevant features jointly over multiple 
timepoints, as well as to obtain coefficients that vary smoothly over time 
seems to result in a clear gain in performance. The Temporal Group LASSO is a 
simple yet powerful extension of the traditional regularized generalized linear 
model concept to longitudinal data, and we believe that this category of models 
deserves more exploration and evaluation in the future.

\bibliographystyle{abbrv}
\bibliography{references}

\end{document}